\documentclass[letterpaper,10pt,conference]{ieeeconf}
\usepackage[T1]{fontenc}
\usepackage{graphicx}
\usepackage{amsmath,amssymb}
\usepackage{bm}
\usepackage{etoolbox}
\usepackage{booktabs,array,tabularx,multirow}
\usepackage[table]{xcolor}
\usepackage{float}
\usepackage{needspace}
\usepackage{stfloats}
\usepackage{placeins}
\usepackage{url}
\usepackage{cleveref}
\crefname{figure}{Fig.}{Figs.}
\crefname{table}{Table}{Tables}
\crefname{section}{Section}{Sections}
\crefname{equation}{Eq.}{Eqs.}

\patchcmd{\abstract}{---}{:}{}{}
\definecolor{ThetaPrimary}{HTML}{5B6CFF}
\definecolor{ThetaTint}{HTML}{F2F4FF}
\newcommand{\bench}{Humanoid\allowbreak ToolBench}
\newcommand{\takeaway}{\noindent\textcolor{ThetaPrimary}{$\boldsymbol{\rightarrow}$}\enspace\textbf{Takeaway.}\ }
\ExplSyntaxOn
\NewDocumentCommand{\scores}{m}
  {
    \tl_set:Nn \l_tmpa_tl {#1}
    \tl_replace_all:Nnn \l_tmpa_tl { / } { \hspace{0.1em}/\hspace{0.1em} }
    \mbox { \tl_use:N \l_tmpa_tl }
  }
\ExplSyntaxOff
\newcommand{\scoresfirst}[1]{\textbf{\scores{#1}}}
\newcommand{\scoressecond}[1]{\underline{\scores{#1}}}

\title{\LARGE\bfseries HumanoidToolBench: Benchmarking Humanoid Tool Use from Selection to Mobile Execution}
\author{Kyochul Jang$^{1}$, Seohyeon Park$^{1}$, Ohchul Kwon$^{1}$, Sangjun Park$^{2}$, Junhyeok Choi$^{1}$, Seungyeop Yi$^{1}$,\\
Chaeyun Kim$^{1}$, Sangkyu Lee$^{1}$, Idan Szpektor$^{3}$, Avi Caciularu$^{3}$, Jongmin Park$^{1}$, and Youngjae Yu$^{1*}$\\[0.3em]
{\normalsize $^{1}$Seoul National University,\quad $^{2}$University of Massachusetts Amherst,\quad $^{3}$Google Research}\\
{\small $^{*}$Corresponding author.}}

\makeatletter
\let\IEEEaftertitletext\@IEEESAVECMDIEEEaftertitletext
\IEEEaftertitletext{%
  \begin{minipage}{\textwidth}
    \centering
    \def\@captype{figure}%
    \includegraphics[width=\linewidth]{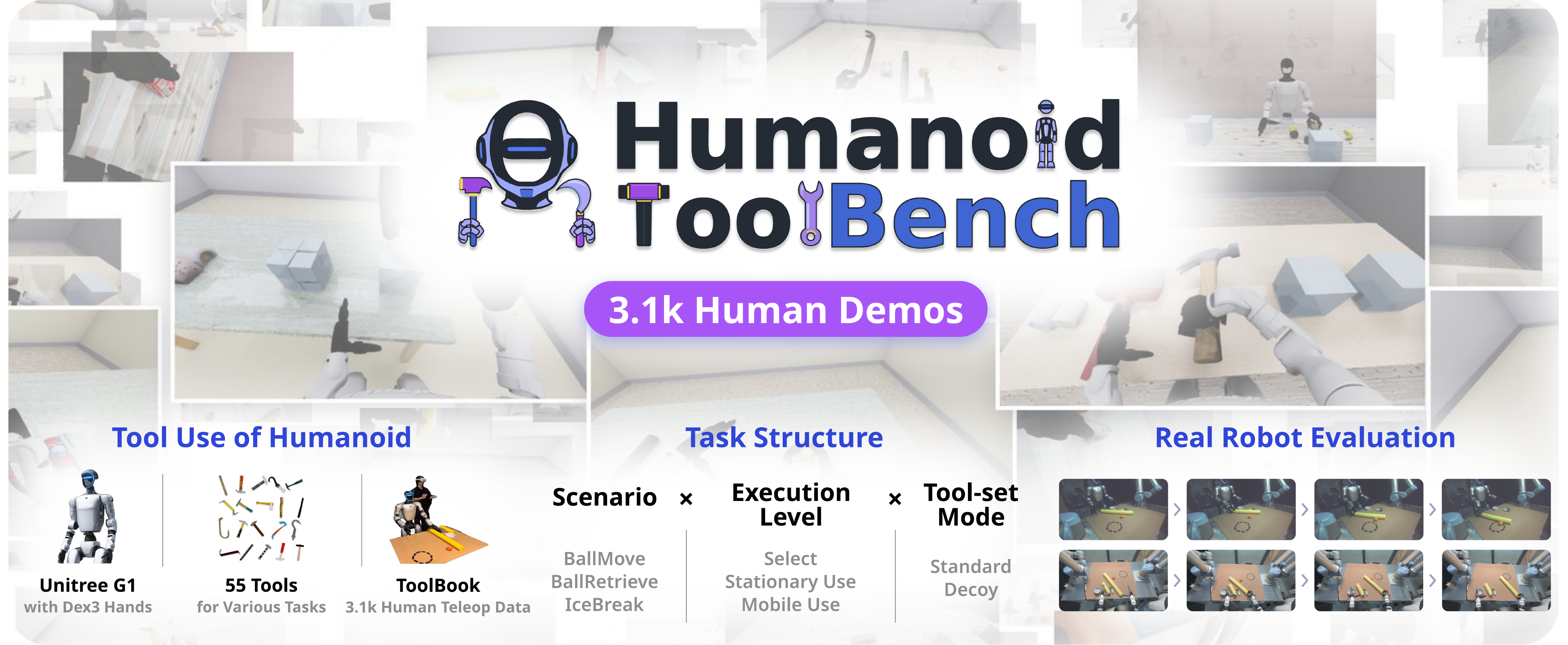}
    \vspace{-0.7cm}
    \caption{\textbf{Overview of \bench{}}. A benchmark for humanoid tool use with 18 tasks spanning three scenarios, three execution levels, and standard and decoy tool sets. With 55 tool assets and 3.1k demonstrations, we evaluate seven policies in simulation and three on the real robot.}
    \label{fig:theta-bench-overview}
  \end{minipage}%
  \vspace{\baselineskip}%
}
\makeatother

\begin{document}
\maketitle
\thispagestyle{empty}
\pagestyle{empty}

\begin{abstract}

As robotic hardware and learning methods advance, humanoids need tools to perform tasks beyond their inherent physical limits. Successful tool use requires selecting a suitable tool and coordinating manipulation and, when needed, locomotion to complete the task. Existing benchmarks do not jointly evaluate these capabilities on a humanoid. We introduce \bench{}, an 18-task benchmark spanning three scenarios, three execution levels, and two tool-set modes, together with ToolBook, a dataset of 3.1k demonstrations collected in simulation and on a real Unitree G1. Evaluation of seven policies in simulation and three on the real robot reveals substantial gaps between selecting a suitable tool and completing the task. Focused GR00T N1.7 probes show reduced selection accuracy on unseen tools and continued task execution under unrelated instructions. Code and data are available at \url{https://snu-pi.github.io/HumanoidToolBench/}.
\end{abstract}

\section{Introduction}
\label{sec:introduction}

Recent advances in robot hardware and learning have accelerated research on humanoids that can perform diverse tasks in human environments~\cite{nvidiaGR00TN1Open2025,sferrazzaHumanoidBenchSimulatedHumanoid2024}. Humans overcome physical limitations by using tools to extend the reach and capabilities of their bodies~\cite{maravitaToolsBodySchema2004,cardinaliToolUseInduces2009}. Humanoids intended to work in the same environments must likewise be able to use tools when their bodies alone cannot accomplish a task. This requires selecting a tool whose properties suit the goal and coordinating grasping, manipulation, and, when needed, locomotion to use it effectively. The grasp determines how much of a tool's reach is available and how its working end is oriented for pushing, pulling, or striking. Body motion must then preserve the spatial relationship needed for contact with the target object. Evaluating humanoid tool use therefore requires testing the full process from selection to task completion.

Existing benchmarks address parts of this problem (\cref{tab:related-work}). HumanoidBench~\cite{sferrazzaHumanoidBenchSimulatedHumanoid2024} includes spoon and window-wiping tasks, but does not evaluate the choice among tools with contrasting functional properties. Dedicated tool-use benchmarks are limited to visual question answering (VQA)~\cite{zhangPhysToolBenchBenchmarkingPhysical2025} or fixed-base arms~\cite{yuanGraspsDexterityLargeScale2026,linRoboWitsUnexpectedChallenges2026}. A benchmark is therefore needed that jointly evaluates functional tool selection and task completion on a humanoid in both stationary and mobile settings.

\begin{table*}[!t]
  \centering
\caption{Comparison of \bench{} with related manipulation, humanoid, and tool-use benchmarks. {\color[HTML]{C99700}$\bm{\triangle}$}, tool use included but not the primary evaluation scope; 0, not applicable or none released; a hyphen, not reported.}
\vspace{-0.2cm}
  \label{tab:related-work}
  \footnotesize
  \setlength{\tabcolsep}{2pt}
  \let\plaincheckmark\checkmark
  \let\plaintimes\times
  \let\plaintriangle\triangle
  \renewcommand{\checkmark}{{\color[HTML]{2E7D32}\large\bm{\plaincheckmark}}}
  \renewcommand{\times}{{\color[HTML]{C62828}\large\bm{\plaintimes}}}
  \renewcommand{\triangle}{{\color[HTML]{C99700}\large\bm{\plaintriangle}}}
  \begin{tabularx}{\textwidth}{
    >{\raggedright\arraybackslash}m{0.15\textwidth}
    >{\centering\arraybackslash}m{0.09\textwidth}
    >{\centering\arraybackslash}m{0.095\textwidth}
    >{\centering\arraybackslash}m{0.07\textwidth}
    >{\centering\arraybackslash}m{0.08\textwidth}
    >{\centering\arraybackslash}m{0.095\textwidth}
    >{\centering\arraybackslash}m{0.105\textwidth}
    >{\centering\arraybackslash}m{0.115\textwidth}
    >{\centering\arraybackslash}m{0.115\textwidth}}
    \toprule
    & & \multicolumn{3}{c}{\textbf{Tool Use}}
      & \multicolumn{2}{c}{\textbf{Task Structure}}
      & \multicolumn{2}{c}{\textbf{\# Human Demos}} \\
    \cmidrule(lr){3-5}\cmidrule(lr){6-7}\cmidrule(lr){8-9}
    \raisebox{0.8ex}{\textbf{Benchmark}}
      & \raisebox{0.8ex}{\textbf{Embodiment}}
      & \shortstack{\textbf{Tool-Use}\\\textbf{Scope}}
      & \shortstack{\textbf{\# tool}\\\textbf{assets}}
      & \textbf{Selection}
      & \shortstack{\textbf{Hierarchical}\\\textbf{Levels}}
      & \shortstack{\textbf{Decoy}\\\textbf{Variants}}
      & \textbf{Sim}
      & \textbf{Real} \\
    \midrule
    LIBERO~\cite{liuLIBEROBenchmarkingKnowledge2023}
      & Single-arm
      & $\times$ & 0 & $\times$ & $\times$ & $\checkmark$
      & 6,500 & 0 \\
    RoboTwin 2.0~\cite{chenRoboTwin20Scalable2026}
      & Dual-arm
      & $\triangle$ & 11 & $\times$ & $\checkmark$ & $\times$
      & 0 & 40 \\
    \addlinespace[1pt]\cmidrule(lr){1-9}\addlinespace[1pt]
    HumanoidBench~\cite{ sferrazzaHumanoidBenchSimulatedHumanoid2024}
      & Humanoid
      & $\triangle$ & $1$ & $\times$ & $\times$ & $\times$
      & 0 & 0 \\
    SIMPLE~\cite{weiSIMPLESimulationBasedPolicy2026}
      & Humanoid
      & $\times$ & 0 & $\times$ & $\checkmark$ & $\times$
      & - & - \\
    WOLF-VLA~\cite{boukheddimiWOLFVLAWholeBodyHumanoid2026}
      & Humanoid
      & $\times$ & 0 & $\times$ & $\checkmark$ & $\times$
      & 0 & 0 \\
    \addlinespace[1pt]\cmidrule(lr){1-9}\addlinespace[1pt]
    PhysToolBench~\cite{zhangPhysToolBenchBenchmarkingPhysical2025}
      & None (VQA)
      & $\checkmark$ & 0 & $\checkmark$ & $\checkmark$ & $\times$
      & 0 & 0 \\
    DexCraft~\cite{yuanGraspsDexterityLargeScale2026}
      & Single-arm
      & $\checkmark$ & 6 & $\times$ & $\times$ & $\times$
      & 30 & 150 \\
    RoboWits~\cite{linRoboWitsUnexpectedChallenges2026}
      & Dual-arm
      & $\checkmark$ & 17 & $\checkmark$ & $\times$ & $\checkmark$
      & 1,207 & 0 \\
    \addlinespace[1pt]\cmidrule(lr){1-9}\addlinespace[1pt]
    \rowcolor{ThetaTint}
    \textbf{Ours}
      & \textbf{Humanoid}
      & $\checkmark$ & \textbf{55} & $\checkmark$ & $\checkmark$ & $\checkmark$
      & \textbf{3,003} & \textbf{91} \\
    \bottomrule
  \end{tabularx}
\end{table*}

To address these limitations, we introduce \bench{}, an 18-task benchmark, and ToolBook, 3.1k human demonstrations collected in simulation and on a real Unitree G1 (\cref{fig:theta-bench-overview}). The design separates selection from execution. Three scenarios, \textit{BallMove}, \textit{BallRetrieve}, and \textit{IceBreak}, require sufficient reach, a shape that can engage and pull, and physical properties that support striking, respectively. Three execution levels distinguish selection and pickup (L0), stationary tool use (L1), and mobile tool use (L2), separating acquisition from task completion. Standard (S) and decoy (D) modes vary the alternatives under the same instruction: standard mode presents the suitable tool among unrelated objects, while decoy mode adds a competing tool that lacks a required property. Combining levels and modes exposes cases where a suitable choice is followed by failed execution, as well as behavior changes induced by competing candidates. ToolBook spans this structure to support policy training and real-robot adaptation.

Our central question is how reliably humanoid policies select and use suitable tools, and how these behaviors respond to changes in tool appearance and task instructions. We examine task completion, selection and execution failures, and responses to changes in tools and instructions. We evaluate seven policies in simulation and three on the real G1, comparing task success across scenarios, execution levels, and modes. Recorded interaction events distinguish contacting a candidate, lifting the suitable tool, and applying it to the target object. They reveal whether low completion accompanies limited pickup or persists despite frequent tool interaction, and capture changes in initial choice that final success can conceal. Focused probes of a separately trained standard-mode GR00T N1.7 policy examine representations and initial contact for unseen suitable tools, and success and motion under instruction changes in a fixed IceBreak scene. The results reveal gaps between acquisition and completion, with further declines when tasks combine locomotion and manipulation. Our contributions are as follows:
\begin{itemize}\setlength{\itemsep}{0pt}
  \item To the best of our knowledge, \bench{} is the first benchmark of humanoid tool use that spans selection and pickup, stationary tool use, and mobile tool use, with ToolBook, a dataset of 3.1k human demonstrations in simulation and on a Unitree G1.
  \item \bench{} combines execution levels, decoy tool sets, and recorded interaction events to distinguish limitations of selection and task execution.
  \item We evaluate seven policies in simulation and three on the real G1 and probes to examine behavior beyond final success.
\end{itemize}

\section{Related Work}
\label{sec:related-work}

\noindent\textbf{Humanoid Benchmarks.} Manipulation benchmarks such as LIBERO~\cite{liuLIBEROBenchmarkingKnowledge2023} and RoboTwin 2.0~\cite{chenRoboTwin20Scalable2026} evaluate vision-language-action models (VLAs) on single- and dual-arm manipulators. Meanwhile, humanoid policies such as GR00T N1.7~\cite{nvidiaGR00TN1Open2025} and $\Psi_0$~\cite{weiPs0OpenFoundation2026} require evaluation on a humanoid, and several benchmarks have been proposed: HumanoidBench~\cite{sferrazzaHumanoidBenchSimulatedHumanoid2024} evaluates whole-body control, including spoon and window-wiping tasks, and SIMPLE~\cite{weiSIMPLESimulationBasedPolicy2026} and WOLF-VLA~\cite{boukheddimiWOLFVLAWholeBodyHumanoid2026} extend evaluation to humanoid VLAs. SIMPLE stages its tasks to report where a policy fails but does not vary the decision a task requires, and WOLF-VLA releases no human demonstrations for training. These benchmarks do not jointly evaluate functional selection, stationary and mobile use, and controlled tool-set variations (\cref{tab:related-work}). Our \bench{} fills this gap: a humanoid selects a tool from 55 assets, with or without a decoy, and uses it while stationary and while moving, with 3,003 simulation and 91 real-robot demonstrations for training.

\noindent\textbf{Tool Use Benchmarks in Robotics.}
Tool use extends the limits of the body~\cite{maravitaToolsBodySchema2004,irikiCodingModifiedBody1996} and is also studied for AI systems~\cite{schickToolformerLanguageModels2023,jangDICEBENCHEvaluatingToolUse2025}, but physical tool use additionally requires a body that grasps, carries, and applies the tool. Three benchmarks target robotic tool use (\cref{tab:related-work}): PhysToolBench~\cite{zhangPhysToolBenchBenchmarkingPhysical2025} evaluates tool understanding through visual question answering without acting on it, DexCraft~\cite{yuanGraspsDexterityLargeScale2026} collects tool-use grasps on a single arm with the tool provided, and RoboWits~\cite{linRoboWitsUnexpectedChallenges2026} asks a dual-arm robot to select a tool in unexpected situations. None evaluates tool use on a humanoid or combines execution levels with controlled tool-set variations: PhysToolBench tiers understanding but not execution, DexCraft and RoboWits do not divide their tasks into progressive levels, and only RoboWits varies the decision a task requires. Our \bench{} evaluates tool use on a humanoid that selects, carries, and applies the tool, and structures every scenario along two axes: execution levels increase the demands of task completion, while standard and decoy modes vary the tool choice under the same instruction. Pairing these axes supports comparisons among policies facing the same alternatives and within a policy as execution demands change. This links a tool's functional suitability to the physical demands of grasping it, contacting the target, and coordinating body motion.

\begin{figure*}[!t]
  \centering
  \includegraphics[width=\textwidth]{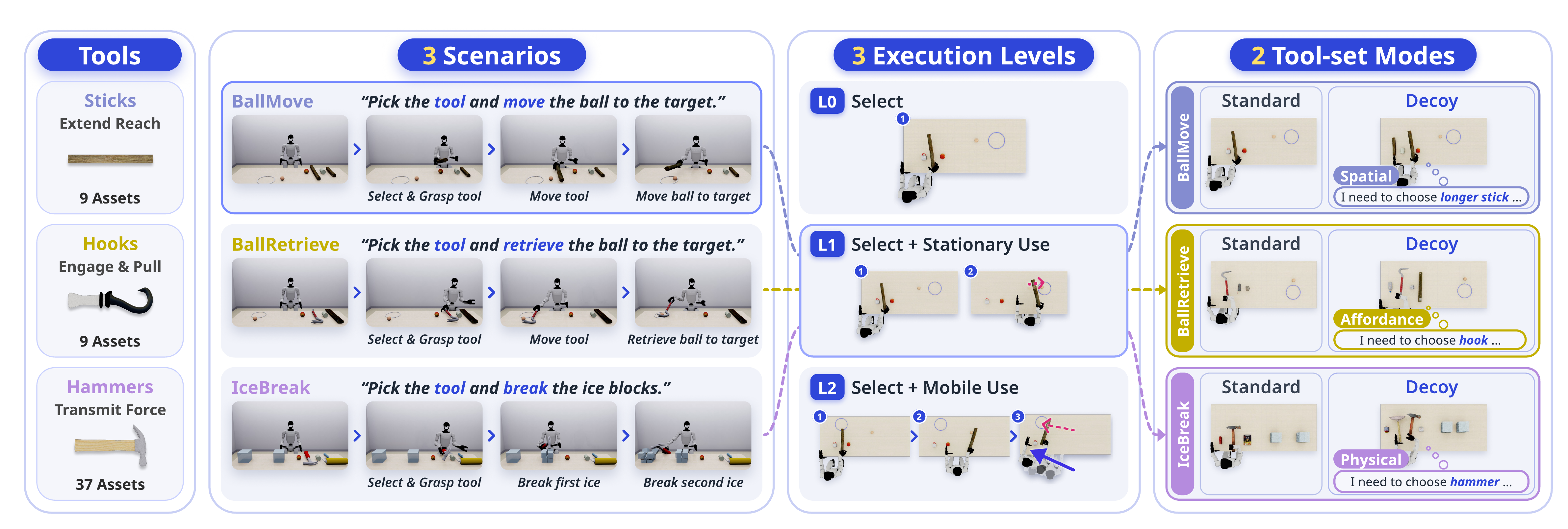}
  \caption{\textbf{Task structure.} 3 scenarios × 3 execution levels × 2 tool-set modes = 18 tasks, built on a library of 55 tool assets spanning three functional categories. Execution levels progress from selection and pickup (L0) to stationary tool use (L1) and mobile tool use (L2). Decoy mode requires choosing the tool that meets the task's spatial, affordance, or physical requirement.}
  \label{fig:task-examples}
\end{figure*}

\Needspace{12\baselineskip}
\section{\bench}
\label{sec:benchmark}

\bench{} evaluates whether humanoid policies select tools suited to the environment and the user's instruction, then use them to complete the requested task (\cref{fig:theta-bench-overview}). We instantiate the hierarchical levels in \cref{tab:related-work} as \emph{execution levels}, which increase the demands of tool use. Each task is a combination of scenario, execution level, and tool-set mode, yielding 18 tasks from three scenarios, three execution levels, and two modes. ToolBook supplies 3.1k human demonstration trajectories from simulation and a real Unitree G1. Sections~\ref{sec:formulation} to \ref{sec:evaluation-protocol} formulate the task and describe its design philosophy, tasks, environment, demonstration collection, and evaluation protocol.

\Needspace{4\baselineskip}
\subsection{Problem Formulation}
\label{sec:formulation}

A humanoid receives a language goal that does not name a tool, and it must select a suitable tool from the candidates in the scene and use it to achieve the goal. We model each task as a goal-conditioned partially observable Markov decision process (POMDP) in which a binary success criterion replaces the reward,
\begin{equation}
  \mathcal{M}=(\mathcal{S},\mathcal{A},\mathcal{O},P,\Omega,G),
  \label{eq:interaction}
\end{equation}
where $\mathcal{S}$ contains robot, object, and controller states, $\mathcal{A}$ and $\mathcal{O}$ are the command and observation spaces, $P(s_{t+1}\mid s_t,a_t)$ combines physics with the fixed whole-body controller, $\Omega(o_t\mid s_t)$ generates observations, and $G$ marks success. Each episode starts with a candidate set $\mathcal{B}$ that holds one suitable tool $b^\star$ and two unrelated objects, plus a decoy tool in decoy mode, and the language goal $g$ describes the task without naming $b^\star$.

The observation $o_t=(I_t,\mathbf{x}_t)$ contains the head-camera image and the robot state in \cref{eq:policy-state}, which carries no object information, so a policy must infer which candidate is suitable from $g$ and the image. Given the history $h_t=(o_0,a_0,\ldots,o_t)$, a policy acts according to
\begin{equation}
  a_t\sim\pi_\theta(\,\cdot\mid h_t,g),
  \label{eq:policy}
\end{equation}
where $\theta$ denotes the parameters of policy $\pi_\theta$, learned from human demonstrations and held fixed during evaluation; an action-chunking policy conditions on an earlier part of $h_t$.

An episode $\tau$ ends when the task succeeds, when it fails, or at the step limit $H$. Its success indicator $G(\tau)\in\{0,1\}$ requires lifting $b^\star$ in selection and pickup tasks and reaching the scenario's goal state in stationary and mobile tool-use tasks. We report the success rate over $N$ attempted episodes, including timeouts and, on the real robot, safety stops,
\begin{equation}
  \hat{J}(\pi_\theta)=\frac{1}{N}\sum_{i=1}^{N}G\big(\tau^{(i)}\big).
  \label{eq:objective}
\end{equation}

\Needspace{4\baselineskip}
\subsection{Design Philosophy and Tasks}
\label{sec:design-philosophy}

\bench{} follows three design principles: real-world utility, interpretability, and functional selection. Each task instantiates the POMDP in \cref{eq:interaction} with a candidate set $\mathcal{B}$, a language goal $g$, and a success criterion $G(\tau)$. For example, $g=$ ``Pick the tool and move the ball to the target'' does not name $b^\star$, so the policy must select it from $\mathcal{B}$.

\noindent\textbf{Real-world utility.} Tool use should be evaluated on functions that tools serve in everyday settings, over the full process from selecting a tool to achieving the user's goal. The \textbf{scenarios} instantiate this principle by covering three common tool functions: extending reach, engaging and pulling, and transmitting force (\cref{fig:task-examples}). Specifically, (i)~\textit{BallMove} requires a long stick to push a ball beyond the robot's arm reach into a 0.12~m circular goal region, imposing a spatial requirement on tool length; (ii)~\textit{BallRetrieve} requires a hook to catch the far side of the ball and pull it into the circular goal region, imposing an affordance requirement on tool shape; and (iii)~\textit{IceBreak} requires the robot to strike two ice blocks hard enough to break them and scatter their fragments, imposing a physical requirement on tool material and structure.

\Needspace{3\baselineskip}
\noindent\textbf{Interpretability.} The evaluation distinguishes selection, pickup, and task completion through execution levels and recorded interaction events. Three execution \textbf{levels} impose progressively greater demands. L0 (selection and pickup) succeeds when $b^\star$ rises at least 0.08~m, evaluating selection and grasping. L1 (stationary tool use) adds the scenario goal, which can be completed without locomotion. L2 (mobile tool use) moves the \textit{BallMove} goal region, the \textit{BallRetrieve} ball, or one \textit{IceBreak} block to the side of the table where the tools lie, so the task calls for moving along the table while holding the tool. Within each episode, recorded events provide a finer account of selection and execution: candidate contact, correct first contact, correct-tool lift, tool-target contact, and task success (\cref{fig:results-visualization}). In these event names, \emph{correct} refers to the suitable tool $b^\star$, and \emph{target} refers to the ball or ice.

\noindent\textbf{Functional selection.} Tool choice and use should respond to the functional requirements of the task rather than to familiar objects and motions. The two tool-set \textbf{modes} test this by varying the alternatives a policy must compare under the same goal $g$. In standard mode (S), $\mathcal{B}$ contains $b^\star$ and two unrelated objects, so recognizing the only tool suffices. Decoy mode (D) adds one decoy tool that differs from $b^\star$ mainly in the property the scenario requires: the same stick at 0.30~m instead of 0.50~m with equal mass in \textit{BallMove}, a straight stick with the hook's length and mass in \textit{BallRetrieve}, and a light, compliant fly swatter, paint roller, or plunger instead of a metal hammer in \textit{IceBreak} (\cref{fig:task-examples}). The instruction and unrelated objects follow the same sampling rules in both modes, while shuffled candidate positions and the added decoy vary the available choices. The scanned tools vary in color and texture as well as task-relevant properties. Section~\ref{sec:tool-use-analysis} compares representations of seen and unseen suitable-tool instances in decoy scenes. Complementary instruction probes vary wording and task relevance while holding the scene fixed.

\subsection{Environment}
\label{sec:environment}

We implement the simulation environment for \bench{} in MuJoCo~\cite{todorovMuJoCoPhysicsEngine2012}. The robot stands in a room in front of a long table, with the candidate objects on the left half of the table and the task objects on the right half. Each episode draws a new scene: it samples $\mathcal{B}$, with tools from 55 Objaverse~\cite{deitkeObjaverseUniverseAnnotated2023} assets curated through MolmoSpaces~\cite{kimMolmoSpacesLargeScaleOpen2026}, shuffles the candidates, and randomizes placements, the robot spawn, surface materials, and lighting.

Simulation and real-robot evaluation use a Unitree G1 with 29 body joints, two seven-joint Dex3 hands, and a head camera that provides the image $I_t$, under the same decoupled control stack adapted from GR00T-WholeBodyControl~\cite{luo2025sonic}. The policy $\pi_\theta$ observes $I_t$ and the state $\mathbf{x}_t$ and outputs the action $a_t$,
\begin{align}
  \mathbf{x}_t &= \big[
  \underbrace{\mathbf{q}^{\mathrm{h}}_t}_{14},
  \underbrace{\mathbf{q}^{\mathrm{a}}_t}_{14},
  \underbrace{\mathbf{q}^{\mathrm{w}}_t}_{3},
  \underbrace{\bar z_t}_{1}\big]\in\mathbb{R}^{32},
  \label{eq:policy-state}\\
  a_t &= \big[
  \underbrace{\mathbf{q}^{\mathrm{h},\star}_t}_{14},
  \underbrace{\mathbf{q}^{\mathrm{a},\star}_t}_{14},
  \underbrace{\mathbf{q}^{\mathrm{w},\star}_t}_{3},
  \underbrace{z_t^\star}_{1},
  \underbrace{\mathbf{u}_t}_{4}\big]\in\mathbb{R}^{36},
  \label{eq:policy-action}
\end{align}
where $\mathrm{h}$, $\mathrm{a}$, and $\mathrm{w}$ index the hand, arm, and waist joints, $\star$ marks targets, $\bar z_t$ and $z_t^\star$ are the previous and new base-height commands, and $\mathbf{u}_t=(v_x,v_y,m,\psi^\star)$ holds forward and lateral velocities, a turning-mode signal, and a target heading. The pretrained GR00T Balance/Walk reinforcement learning controllers execute the locomotion components of $a_t$, producing the lower body targets required for bipedal motion.

\par\medskip\noindent\begin{minipage}{\columnwidth}
\subsection{Demonstration Collection}
\label{sec:demonstrations}

ToolBook records human tool selection and use through Meta Quest teleoperation in simulation and on the real G1. Eight experienced teleoperators made 1,959 simulation attempts at L1 and L2, yielding 100 successful demonstrations for each of the 12 L1/L2 tasks, 1,200 in total. For every attempt in which $b^\star$ rises 0.08~m, we truncate the trajectory at the first such frame, yielding 1,803 L0 trajectories without separate collection. These comprise 1,189 from the successes, since the other 11 used $b^\star$ without lifting it that high, and 614 from unsuccessful attempts. With 91 curated real-robot demonstrations of L1 \textit{BallMove} and \textit{BallRetrieve}, the resulting ToolBook dataset comprises 3,094 trajectories, with the following breakdown: \[ \underbrace{1{,}803}_{\text{simulation L0}} + \underbrace{1{,}200}_{\text{simulation L1/L2}} + \underbrace{91}_{\text{real-robot L1}} = \underbrace{3{,}094}_{\text{total trajectories}} \] Refer to \cref{tab:dataset-coverage} for the counts by task.

\begin{table}[H]
  \centering
  \caption{ToolBook coverage by task.}
  \label{tab:dataset-coverage}
  \footnotesize
  \setlength{\tabcolsep}{4pt}
  \renewcommand{\arraystretch}{1.1}
  \begin{tabularx}{\columnwidth}{l*{3}{>{\centering\arraybackslash}X}}
    \toprule
    \rowcolor{ThetaTint}
    \multicolumn{4}{l}{\textbf{Simulation: 3,003 trajectories}} \\
    Scenario & L0 (S/D) & L1 (S/D) & L2 (S/D) \\
    \midrule
    \textit{BallMove} & 289/244 & 100/100 & 100/100 \\
    \textit{BallRetrieve} & 388/313 & 100/100 & 100/100 \\
    \textit{IceBreak} & 316/253 & 100/100 & 100/100 \\
    \midrule
    \rowcolor{ThetaTint}
    \multicolumn{4}{l}{\textbf{Real robot: 91 trajectories}} \\
    Scenario & Execution level & S & D \\
    \midrule
    \textit{BallMove} & L1 & 29 & 21 \\
    \textit{BallRetrieve} & L1 & 20 & 21 \\
    \bottomrule
  \end{tabularx}
  \par\smallskip
  \begin{minipage}{\columnwidth}
    \footnotesize
    Only collected tasks are shown. S/D: standard/decoy counts.
    Simulation counts are extracted L0 prefixes and successful L1/L2 episodes;
    real-robot counts are curated demonstrations.
  \end{minipage}
\end{table}
\end{minipage}\par

\subsection{Evaluation Protocol}
\label{sec:evaluation-protocol}

Policies for simulation evaluation are trained on simulation demonstrations (sim-only training); those for real evaluation are additionally fine-tuned on real-robot demonstrations (sim+real training). Simulation evaluation covers all 18 tasks, and real-robot evaluation covers L1 \textit{BallMove} and \textit{BallRetrieve} in both modes. We report the success rate $\hat{J}(\pi_\theta)$ in \cref{eq:objective} as a percentage; a simulated episode succeeds once its criterion in Section~\ref{sec:design-philosophy} holds for one second. Rates are reported by scenario, level, and mode within each evaluation domain.

\section{Experiments}
\label{sec:results}

\begin{table*}[!t]
  \centering
  \caption{Simulation success rate (\%) by scenario, execution level, and tool-set mode. Within each cell, values are reported in the order BallMove / BallRetrieve / IceBreak. Rows in each family are ordered by mean L1/L2 success. In each column, bold and underlined cells have the highest and second-highest mean over the three scenarios.}
  \label{tab:baseline-results}
  \footnotesize
  \setlength{\tabcolsep}{3pt}
  \begin{tabularx}{\textwidth}{@{}ll*{6}{>{\centering\arraybackslash}X}@{}}
    \toprule
    \rowcolor{ThetaTint} & & \multicolumn{3}{c}{Standard (S)} & \multicolumn{3}{c}{Decoy (D)} \\
    \cmidrule(lr){3-5}\cmidrule(lr){6-8}
    Family & Policy & L0 & L1 & L2 & L0 & L1 & L2 \\
    \midrule
    \multirow{2}{*}{IL} & ACT~\cite{zhaoLearningFineGrainedBimanual2023} & \scores{5/0/0} & \scores{0/0/0} & \scores{0/0/0} & \scores{3/4/1} & \scores{0/0/0} & \scores{0/0/0} \\
     & DP~\cite{chiDiffusionPolicyVisuomotor2023} & \scores{14/5/17} & \scores{0/0/2} & \scores{0/0/0} & \scores{11/9/13} & \scores{0/0/1} & \scores{0/0/0} \\
    \midrule
    \multirow{3}{*}{VLA} & $\pi_{0.5}$~\cite{black$pi_05$VisionLanguageActionModel2025} & \scores{29/24/31} & \scores{4/0/12} & \scores{1/0/0} & \scores{26/28/21} & \scores{5/3/1} & \scores{1/1/0} \\
     & $\Psi_0$~\cite{weiPs0OpenFoundation2026} & \scores{19/14/22} & \scores{9/0/13} & \scores{1/0/2} & \scores{14/15/18} & \scores{7/3/8} & \scores{3/1/0} \\
     & GR00T N1.7~\cite{nvidiaGR00TN1Open2025} & \scoresfirst{77/73/88} & \scoressecond{64/36/66} & \scoressecond{10/9/31} & \scoressecond{67/59/75} & \scoressecond{58/30/63} & \scoresfirst{30/5/48} \\
    \midrule
    \multirow{2}{*}{WAM} & Cosmos 3 Edge Policy~\cite{nvidiaCosmos3Omnimodal2026} & \scores{12/0/2} & \scores{0/0/1} & \scores{0/0/0} & \scores{14/0/0} & \scores{2/0/0} & \scores{1/0/0} \\
     & FastWAM~\cite{yuanFastWAMWorldAction2026} & \scoressecond{64/62/98} & \scoresfirst{76/57/86} & \scoresfirst{9/20/69} & \scoresfirst{74/45/93} & \scoresfirst{75/45/82} & \scoressecond{8/11/61} \\
    \bottomrule
  \end{tabularx}
\end{table*}

\begin{table}[!t]
  \centering
  \caption{Real-robot L1 success (successes / 10 trials). Policies are trained on simulation demonstrations and fine-tuned on real-robot demonstrations (sim+real).}
  \label{tab:real-results}
  \footnotesize
  \setlength{\tabcolsep}{3pt}
  \begin{tabularx}{\linewidth}{@{}l*{4}{>{\centering\arraybackslash}X}@{}}
    \toprule
    \rowcolor{ThetaTint} & \multicolumn{2}{c}{\textit{BallMove}} & \multicolumn{2}{c}{\textit{BallRetrieve}} \\
    \cmidrule(lr){2-3}\cmidrule(lr){4-5}
    Policy & S & D & S & D \\
    \midrule
    ACT     & 1/10          & 0/10          & 2/10          & 1/10 \\
    GR00T N1.7   & \textbf{6/10} & \textbf{4/10} & \textbf{4/10} & 4/10 \\
    FastWAM & 5/10          & \textbf{4/10} & \textbf{4/10} & \textbf{5/10} \\
    \bottomrule
  \end{tabularx}
  \par\medskip
  \makeatletter\def\@captype{figure}\makeatother
  \includegraphics[width=\linewidth]{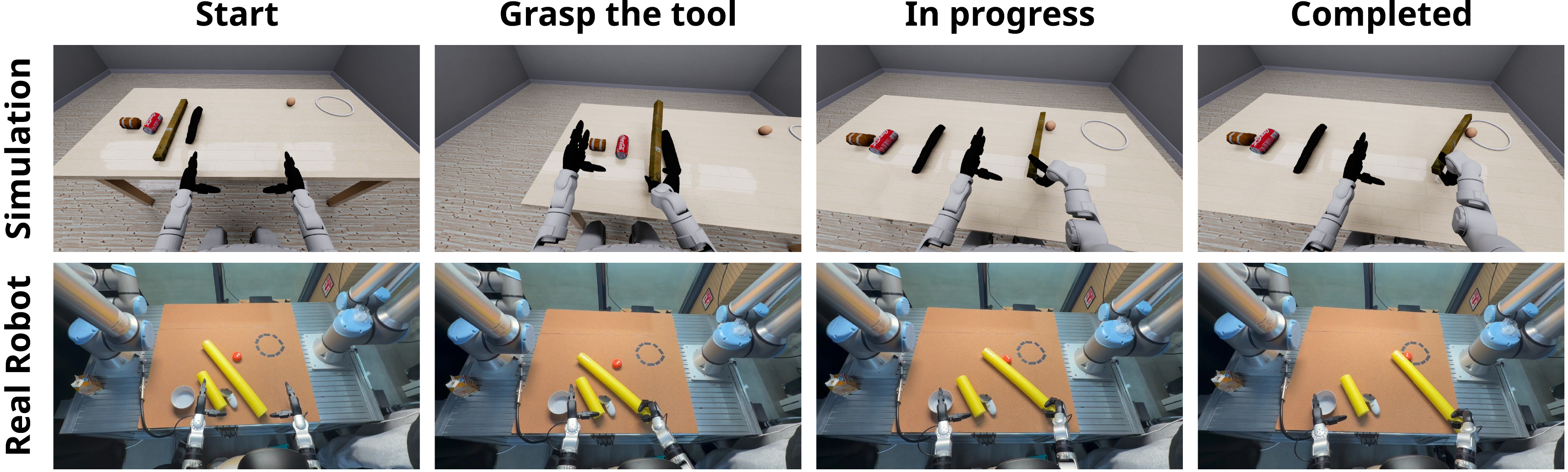}
  \caption{GR00T N1.7 \textit{BallMove} rollouts in simulation and on the real robot at four matched phases.}
  \label{fig:sim-real-rollout}
\end{table}

We first compare task success in simulation and on the real G1 and inspect rollout failures (Section~\ref{sec:benchmark-results}). We then examine execution events to locate the gaps between contacting a candidate, lifting the suitable tool, and completing the task (Section~\ref{sec:selection-analysis}). Finally, we examine GR00T N1.7's tool-region representations and successful motions under tool and instruction changes (Section~\ref{sec:tool-use-analysis}).

\subsection{Experimental Setup}
\label{sec:experimental-setup}

\noindent\textbf{Policies.} We compare three VLAs, GR00T N1.7 (3.14B)~\cite{nvidiaGR00TN1Open2025}, $\Psi_0$ (2.63B)~\cite{weiPs0OpenFoundation2026}, and $\pi_{0.5}$ (3.62B)~\cite{black$pi_05$VisionLanguageActionModel2025}, two world-action models (WAMs), Cosmos 3 Edge Policy (3.37B)~\cite{nvidiaCosmos3Omnimodal2026} and FastWAM (6.02B)~\cite{yuanFastWAMWorldAction2026}, and two imitation-learning (IL) baselines, ACT (52.1M)~\cite{zhaoLearningFineGrainedBimanual2023} and DP (84.7M)~\cite{chiDiffusionPolicyVisuomotor2023}. Sizes are rounded parameter counts for policy checkpoints, including integrated frozen backbones but excluding external frozen encoders and VAEs; tied weights are counted once. These policies cover different approaches to learning robot behavior; we assess their ability to select and use tools, with the IL baselines providing task-specific imitation references.

\noindent\textbf{Training.} For simulation evaluation, every policy learns from the 3,003 simulation trajectories in ToolBook (sim-only training). The policies use DROID checkpoints for GR00T N1.7, $\pi_{0.5}$, and Cosmos 3, and model-specific pretrained weights for $\Psi_0$ and FastWAM. ACT and DP gain language conditioning from a frozen CLIP ViT-L/14 text encoder~\cite{radfordLearningTransferableVisual2021}. Each policy is trained once for 40,000 updates with 128 samples per update, drawing at least one sample from each of the 18 tasks. For real evaluation, ACT, GR00T N1.7, and FastWAM continue from their simulation-trained weights for 5,000 updates on the 91 real-robot demonstrations (sim+real training).

\noindent\textbf{Evaluation.} In simulation, we run 100 episodes for each of the 18 tasks, with initial seeds matched across policies and a one-minute limit of $H=3{,}000$ control steps at 50~Hz. In the real setting, ACT, GR00T N1.7, and FastWAM are evaluated on the Unitree G1 in L1 \textit{BallMove} and \textit{BallRetrieve}, each in S and D modes, with 10 trials per scenario-mode combination and a one-minute limit per trial. \Cref{fig:sim-real-rollout} shows example GR00T N1.7 \textit{BallMove} rollouts in simulation and on the real robot at matched phases. Every policy receives the same observation, the left head-camera image and the robot state $\mathbf{x}_t$, and executes each predicted action chunk open loop before the next query.
Each policy uses the chunk length it was trained with: 100 actions (2~s) for ACT, 40 for GR00T N1.7, 32 for FastWAM, 30 for $\Psi_0$, and 16 for DP, $\pi_{0.5}$, and Cosmos 3 Edge Policy. ACT and DP condition on a single observation, and no policy uses temporal ensembling.

\subsection{Results}
\label{sec:benchmark-results}

\cref{tab:baseline-results} compares all seven policies in simulation, and \cref{tab:real-results} evaluates three policies on the real G1 after sim+real training. Together, they show how completion varies across policies, execution demands, and candidate sets, and whether the broad performance pattern persists on the real robot.

\noindent\textbf{Stationary tool use depends on both the policy and tool function.} ACT, DP, and Cosmos 3 Edge Policy rarely complete L1 tasks, while $\pi_{0.5}$ and $\Psi_0$ achieve modest success, particularly in \textit{IceBreak}. GR00T N1.7 and FastWAM perform substantially better across all three scenarios in both modes, with FastWAM attaining the highest L1 success throughout. Even these stronger policies perform less well at retrieving than at pushing or striking, indicating that capability varies across tool functions. L0 separately evaluates selection and pickup under its own success criterion: FastWAM, for example, achieves 64\% at L0 and 76\% at L1 on standard \textit{BallMove}. These are success rates in distinct task conditions. To locate failures between acquisition and completion, Section~\ref{sec:selection-analysis} examines recorded interaction events within the same L1/L2 evaluation rollouts.

\begin{figure}[t]
  \vspace*{3mm}
  \centering
  \includegraphics[width=\columnwidth]{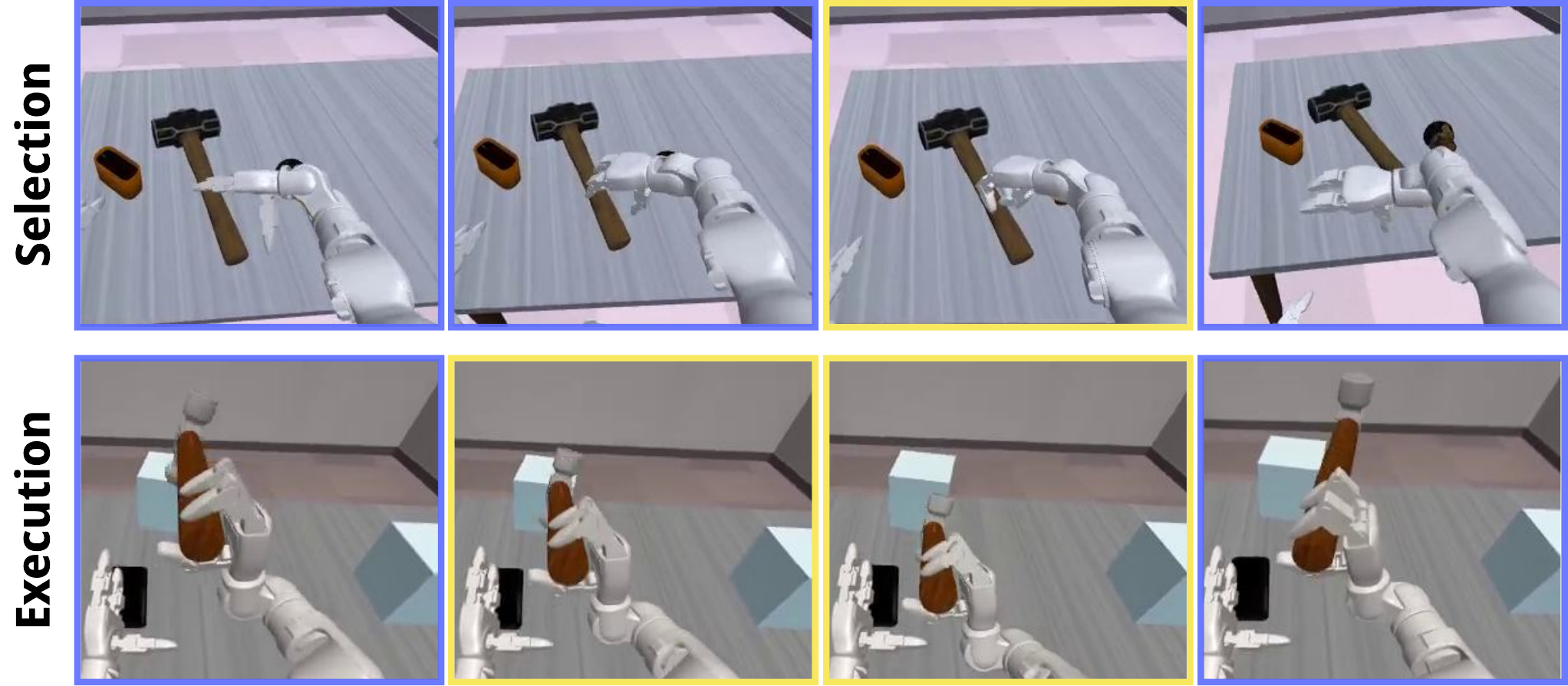}
  \caption{\textbf{Selection and spatial alignment errors.} Reviewed rollouts show decoy pickup (top) and tool-target alignment errors (bottom). Yellow borders mark the frames where spatial alignment errors occur.}
  \label{fig:rollout-errors}
\end{figure}

\begin{figure}[t]
\centering
  \includegraphics[width=\linewidth]{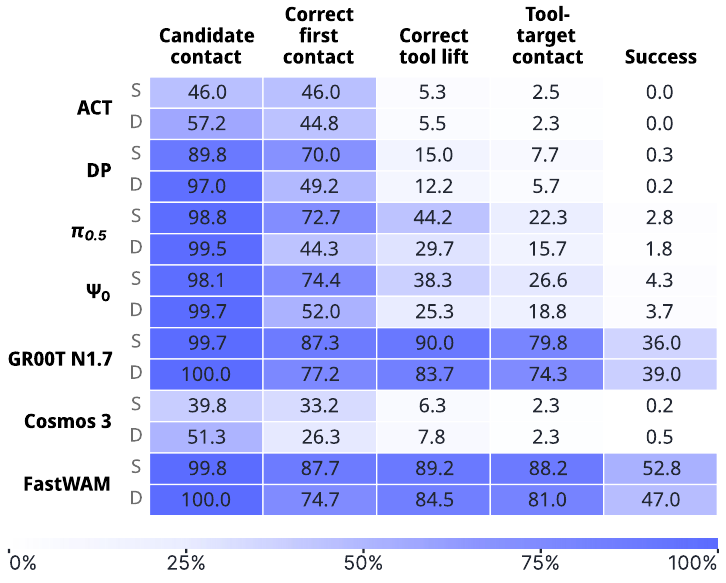}
  \caption{Execution-event success rate (\%) by policy in S and D modes, averaged equally over the three scenarios and over L1 and L2. Success requires the goal state to hold for one second (Section~\ref{sec:evaluation-protocol}).}
  \label{fig:results-visualization}
\end{figure}

\begin{figure}[t]
\centering
  \includegraphics[width=\columnwidth]{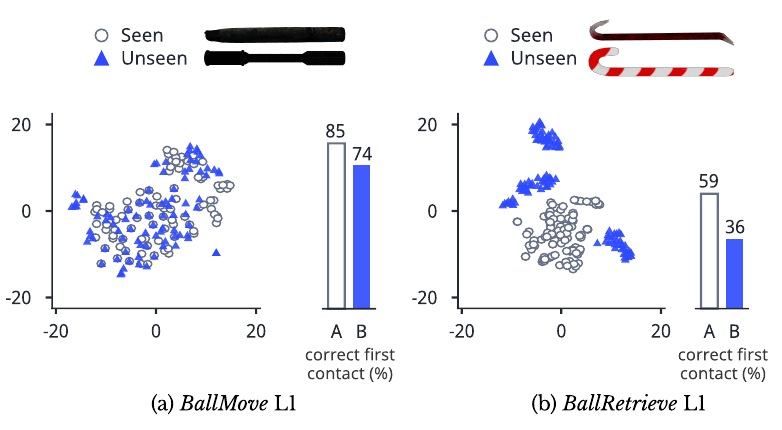}
  \caption{\textbf{Tool-instance changes in L1 decoy scenes.} GR00T N1.7 tool-region t-SNE projections and correct first contact for (a) BallMove and (b) BallRetrieve. Open circles denote seen tools; triangles denote unseen tools. Bars A and B give correct first contact for seen and unseen tools, respectively, over 100 episodes each.}
  \label{fig:tool-transfer}
\end{figure}

\noindent\textbf{Mobile tool use challenges even strong stationary policies.} For every policy, success decreases from L1 to L2 whenever L1 success is nonzero. Thus, stationary tool-use capability does not extend reliably to tasks that also require moving the body. The decline is especially pronounced for FastWAM on \textit{BallMove}, where standard-mode success falls from 76\% to 9\%; GR00T N1.7 also loses much of its stationary performance on both ball tasks. The limited L1 success of $\pi_{0.5}$ and $\Psi_0$ largely disappears at L2, while ACT, DP, and Cosmos 3 Edge Policy already struggle at L1. For GR00T N1.7 and FastWAM, \textit{IceBreak} retains more success than either ball task, indicating that the mobile-use difficulty depends on the required interaction. The L1-to-L2 decline identifies mobile tool use as a major challenge, with difficulty varying across the required interactions.

\noindent\textbf{Decoys challenge strong stationary policies.} GR00T N1.7 and FastWAM score lower in D mode across all three L1 scenarios. For FastWAM, \textit{BallRetrieve} success drops from 57\% to 45\%, despite its relatively strong stationary tool use. This consistent L1 decline shows that competing candidates challenge performance even without locomotion. Effects at L2 are mixed: GR00T N1.7 improves from 10\% to 30\% on \textit{BallMove}, while FastWAM declines across all three scenarios. These outcomes connect task completion to the functional demands of choosing among tools. Section~\ref{sec:selection-analysis} examines how competing candidates change initial choice, suitable-tool pickup, and subsequent tool-target contact.

\noindent\textbf{Stronger simulated policies also lead on the real robot.} GR00T N1.7 and FastWAM outperform ACT in every real-robot condition, consistent with their stronger simulated task completion. For example, on standard \textit{BallMove}, they succeed in 6/10 and 5/10 trials, respectively, compared with 1/10 for ACT. This performance follows only 5,000 fine-tuning updates from the sim-trained policies using 91 real-robot demonstrations. Thus, policies that learn effective tool use in simulation also achieve substantial completion on the real G1 after a short adaptation stage with a small real-data budget. ACT remains less successful after the same adaptation procedure.

\noindent\textbf{Rollout inspection reveals selection and spatial alignment errors.} Visual inspection of GR00T N1.7 and FastWAM evaluation videos identifies two failure patterns (\cref{fig:rollout-errors}). In decoy scenes, both models sometimes grasp and lift the decoy successfully, showing a mismatch between the selected tool and the task despite successful pickup. Other reviewed failures involve grasp poses and tool orientations that prevent effective contact with the target object. These cases point to difficulties in reasoning about tool geometry, object distance, and the orientation needed for the intended interaction. The observations identify task-dependent selection and spatial alignment as challenges beyond acquiring a tool.

\begin{samepage}
\takeaway Tool-use performance varies across policies and tool functions. Mobile tasks remain challenging, and decoys reduce stationary success for the strongest policies. Stronger simulated performance persists after limited real-data adaptation.
\par\end{samepage}

\subsection{Where Do Policies Fail in Tool Use?}
\label{sec:selection-analysis}

\Cref{fig:results-visualization} reports candidate contact, correct first contact, correct-tool lift, tool-target contact, and success. Each value equally averages the event rates over six conditions: three scenarios at L1 and L2 within each mode. L0 is excluded. Each rate measures the fraction of episodes containing the event; an episode can include an incorrect first contact followed by a suitable-tool lift.

\noindent\textbf{High contact rates need not yield pickup or completion.} In standard mode, DP and $\pi_{0.5}$ contact a candidate in 89.8\% and 98.8\% of episodes but lift the suitable tool in only 15.0\% and 44.2\%. GR00T N1.7 and FastWAM have higher lift rates, 90.0\% and 89.2\%, and tool-target contact rates of 79.8\% and 88.2\%, while success remains at 36.0\% and 52.8\%. These contrasts distinguish limited pickup from frequent tool interaction accompanied by lower task success.

\noindent\textbf{Decoys affect first contact more than final success.} For GR00T N1.7 and FastWAM, correct first contact falls from 87.3\% to 77.2\% and from 87.7\% to 74.7\% between standard and decoy modes. Success changes less, from 36.0\% to 39.0\% and from 52.8\% to 47.0\%. In decoy mode, correct-tool lift remains more frequent than correct first contact, at 83.7\% and 84.5\%, showing that initial choice and eventual pickup need not coincide.

\takeaway Weaker policies struggle to lift the suitable tool, while stronger policies often contact the target without completing the task. For GR00T N1.7 and FastWAM, decoys affect initial choice more than final success.

\subsection{Responses to Tool and Instruction Changes}
\label{sec:tool-use-analysis}

\begin{figure}[H]
\centering
  \includegraphics[width=\columnwidth]{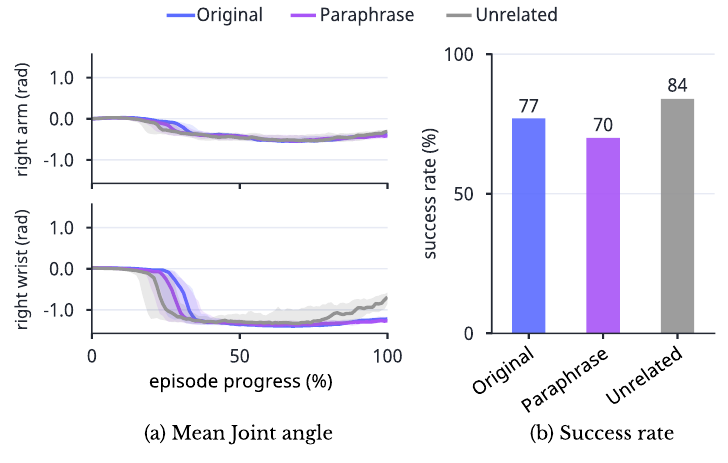}
  \caption{\textbf{Instruction changes in IceBreak L1.} GR00T N1.7 under original, paraphrased, and unrelated instructions in one fixed standard-mode scene. (a) Per-rollout mean angle of the seven right-arm joints, shoulder to wrist (top), and of the three right-wrist joints (bottom). Lines show medians across successful rollouts at normalized episode progress, with interquartile bands. (b) Task success over 100 rollouts per instruction.}
  \label{fig:instruction-motion}
\end{figure}

For these probes, we use GR00T N1.7 trained on 1,593 standard-mode demonstrations for 40,000 updates. \Cref{fig:tool-transfer} compares seen and unseen suitable-tool instances in paired L1 decoy scenes; \cref{fig:instruction-motion} varies instructions in a fixed L1 standard \textit{IceBreak} scene. Seen instances appeared during training, while unseen instances meet the same task requirement.

\Needspace{3\baselineskip}
\noindent\textbf{Tool replacement changes representations and initial contact.} \Cref{fig:tool-transfer} shows t-SNE projections of initial post-normalization visual features pooled over the suitable-tool region, using 91 paired scenes for \textit{BallMove} and 100 for \textit{BallRetrieve}. Seen and unseen features substantially overlap in \textit{BallMove}, whereas unseen instances form groups apart from most seen instances in \textit{BallRetrieve}. Correct first contact decreases from 85\% to 74\% in \textit{BallMove} and from 59\% to 36\% in \textit{BallRetrieve}, measured over 100 episodes per condition. The illustrated hooks differ in shape and surface appearance, and \textit{BallRetrieve} exhibits both clearer feature separation and the larger selection drop. Despite preserved task suitability, these observations suggest limited generalization to out-of-distribution tool appearances.

\noindent\textbf{High task success persists under unrelated instructions.} In the fixed \textit{IceBreak} scene, GR00T N1.7 achieves 77\% success with the original instruction, 70\% with a paraphrase, and 84\% with an unrelated sentence (\cref{fig:instruction-motion}). The prompts are ``Pick the tool and break the ice blocks,'' ``Break the ice blocks using a tool you pick,'' and ``The weather is nice today,'' respectively. Removing the task-relevant request therefore leaves the policy frequently executing and completing the familiar task. The motion profiles support this observation: across successful rollouts, all three conditions show similar broad right-arm profiles and a shared middle-phase wrist pattern. We examine these joints because the arm and wrist position and orient the tool-holding hand. The unrelated condition starts the wrist-angle decrease earlier and returns further toward the initial angle near completion, while retaining the overall task behavior. Because each training scene determines its task, this result shows that the policy completes the task without relying on the instruction; testing instruction following requires conditions in which the instruction conflicts with the scene.

\takeaway GR00T N1.7 selects unseen suitable tools less reliably and continues the familiar \textit{IceBreak} task under unrelated language. These probes show limited transfer across tool appearances and suggest weak instruction following despite high task success.

\section{Discussion and Limitations}
\label{sec:discussion}

The main benchmark comparison uses one training run per policy, with real-robot evaluation of three policies on two L1 scenarios and focused GR00T N1.7 probes. Episode variation reflects scene and rollout variability, rather than training-seed uncertainty. Execution levels vary in layout as well as locomotion demands, and the real-robot study evaluates simulation training followed by real-data adaptation as a complete pipeline. First contact serves as a proxy for initial choice. Tool-region projections characterize visual-feature organization; establishing functional recognition requires controls for appearance. The instruction probe uses one fixed IceBreak scene and varies wording and task relevance; testing instructions for different goals would further assess goal switching. Broader tools, scenes, goals, and additional training seeds would extend the scope of these findings.

The rollout inspection highlights spatial alignment difficulties during tool acquisition and use. Egocentric visual input may contribute by providing limited context for estimating tool-object distance and relative orientation. This explanation is a hypothesis motivated by the reviewed videos. Controlled comparisons of camera viewpoints and spatial observations would clarify the contribution of visual context, alongside the policy's representation, grasping, and control. Similarly, separate simulation-only and real-data adaptation comparisons would measure the contribution of fine-tuning to real-robot performance.

\section{Conclusion}
\label{sec:conclusion}

\bench{} and ToolBook support evaluating humanoid tool use from selection and pickup to mobile task completion. Across seven policies, execution events distinguish failures to lift a suitable tool from failures to complete the task after contact, while decoys change initial choices more than final success. Focused probes additionally reveal sensitivity to unseen tool appearances and continued task execution under unrelated language. These findings motivate policies that adapt selection and execution to functional requirements and task goals beyond familiar training patterns.

\bibliographystyle{IEEEtran}
\bibliography{references}

\end{document}